\documentclass{article} 
\usepackage[final]{colm2026_conference}

\usepackage{microtype}
\usepackage{hyperref}
\usepackage{url}
\usepackage{booktabs}
\usepackage{amsmath}
\usepackage{wrapfig}
\usepackage{graphicx}
\usepackage[most]{tcolorbox}

\usepackage[pass]{geometry}

\usepackage{lineno}

\definecolor{darkblue}{rgb}{0, 0, 0.5}
\hypersetup{colorlinks=true, citecolor=darkblue, linkcolor=darkblue, urlcolor=darkblue}

\title{PM-Bench: Evaluating Prospective Memory in LLM Agents}

\author{Genglin Liu \\
University of California, Los Angeles \\
\texttt{genglinliu@cs.ucla.edu} \\
\And
Saadia Gabriel \\
University of California, Los Angeles \\
}

\begin{document}

\ifcolmsubmission
\linenumbers
\fi

\maketitle

\begin{abstract}
A significant challenge in agentic AI is prospective memory: the ability to execute an intention at a specific future cue or state while other activities are ongoing. We introduce PM-Bench, a text-based benchmark for measuring prospective memory capabilities in modern LLM agents. Inspired by the Virtual Week paradigm from cognitive science, PM-Bench evaluates how well LLM agents maintain user intentions, execute delayed intentions, and monitor latent environment changes. Over the course of a simulated seven-day week, agents must continue an ongoing activity while deciding whether any deferred task is due. We compare eight state-of-the-art LLMs on PM-Bench under eight different agent configurations. PM-Bench proves challenging across all settings: the best method, a GPT-5.4 agent, reaches only 65.1\% F1 score under our evaluation. Furthermore, no single strategy for improving prospective memory dominates across models. We release PM-Bench as a controlled testbed for diagnosing these failures and developing training or inference-time interventions that support reliable prospective behavior \footnote{Benchmark repo at \url{https://github.com/genglinliu/PMBench.git}}.

\end{abstract}

\section{Introduction}

\begin{figure*}[!t]
  \centering
  \includegraphics[width=\textwidth]{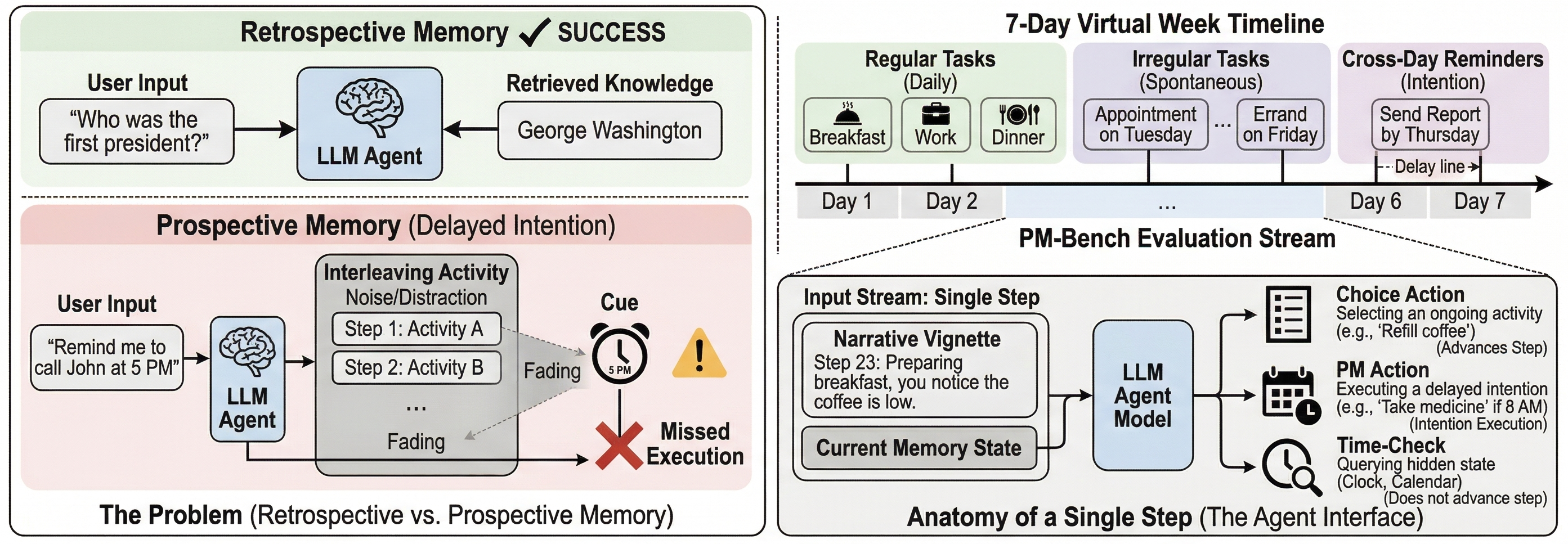}
  \caption{Overview of PM-Bench. The benchmark adapts the \textit{Virtual Week} paradigm into a seven-day text-based schedule in which an LLM agent must carry out deferred intentions at the correct future time or event while continuing an ongoing activity. We propose a series of competitive baselines that attempt to solve PM-Bench and expose critical challenges faced by different agentic setups.}
  \label{fig:fig1}
\end{figure*}

LLM-driven agentic systems are increasingly expected to sustain long-horizon tasks, as models can now browse, code, plan, use tools, and act over many turns rather than a single prompt \citep{cao2026qwen3,team2025kimi,wang2024openhands,jimenez2023swe,zhang2026deepplanning}. As LLM assistants grow more capable, failures often occur not because the system lacks relevant knowledge, but because it does not act on that knowledge at the right moment \citep{gonzalez2025robotouille, liu2025webcoach}. As illustrated in Figure~\ref{fig:fig1}, a reliable assistant, in principle, should remember to carry out tasks that were mentioned at a previous time. It should track rescheduling or cancellations, and stop acting on plans that have already been completed or fulfilled. In cognitive science, this capability is categorized as prospective memory: remembering to perform delayed intentions at an appropriate time or event in the future \citep{mcdaniel2007prospective,brandimonte2014prospective}. Prospective memory is distinct from retrospective recall because success depends not only on what is remembered, but also on whether tasks can be completed under distraction, interference, and monitoring demands \citep{smith2005effects,dismukes2006concurrent,anderson2018proceeding}. Its importance is not merely theoretical: in aviation, failures to remember deferred cockpit actions have been linked to a substantial share of major airline accidents, and in healthcare, self-reported forgetting of blood pressure medication has been associated with worse cardiovascular outcomes \citep{dismukes2006concurrent,nelson2006self}. Classic work on human forgetting argues that prospective memory failures may account for roughly 50\% of these lapses in our daily life \citep{crovitz1984measurements}.

This distinction matters for agentic benchmarking. Existing long-horizon evaluations for LLMs increasingly stress software engineering, browsing, planning, proactive interaction, and memory over extended sessions \citep{deng2023prompting,castillo2024beyond,liu2025proactiveeval,jimenez2023swe,wei2025browsecomp,zhang2026deepplanning,bian2026realmem}. Agent memory methods and benchmarks have focused on retaining past user information, context engineering that's essentially implemented by retrieving from prior episodes or experiences, or extending memory capacity through external stores and architectural changes \citep{li2025toward,lee2025realtalk,packer2023memgpt,zhong2024memorybank,wang2026memex,cheng2026conditional,zhang2025recursive}. However, none of these existing efforts cleanly isolate whether an intention is executed at the correct future cue or time. Recent concurrent work also suggests that current LLMs continue to struggle with this problem \citep{mittal2026did}.

In light of this observation, we introduce PM-Bench, a text-based benchmark for prospective memory in LLM agents. PM-Bench adapts the classic \textit{Virtual Week} paradigm from cognitive science to a format that is executable for modern tool-using LLM agents \citep{rendell2000virtual,rendell2009review,mioni2015test}. In PM-Bench, scenarios unfold over a seven-day schedule. At every step, the agent must continue an ongoing activity while deciding whether any deferred intention should be executed at that given moment. A core design goal of PM-Bench is to separate \emph{what} the agent remembers from \emph{when} it acts. The benchmark consists of event-based and time-based tasks, cross-day obligations, rescheduling and cancellations, as well as invisible state channels that require active monitoring. The design principles of PM-Bench are inspired by relevant psychology studies on human subjects \citep{einstein1990normal,kourtesis2023ecologically}.

PM-Bench comes with a comprehensive scoring pipeline to track fine-grained execution errors, including misses, late completions, update or dependency violations, as well as monitoring overhead. We propose a number of baselines for PM-Bench, with both single-agent and multi-agent strategies. Across these settings, the main pattern is not simply that models ``forget''; it is that prospective behavior is hard to calibrate. Methods that recover more due tasks often do so by becoming over-aggressive and increasing false alarms, whereas conservative methods preserve precision by under-acting. PM-Bench is designed to make this gap visible and analyzable, providing a controlled testbed for future training, inference-time scaffolding, and agent-memory interventions.

In summary, this paper makes the following contributions:
\begin{itemize}
  \item We introduce PM-Bench, a benchmark that evaluates LLM agents' prospective memory under ongoing activity. To this end, we provide a synthetic scenario framework with scalable solvability validation, supporting cross-day tasks, environment monitoring, and dynamic task updates (\S\ref{sec:scenario-design}--\ref{sec:solvability-validation}).
  \item We define an evaluation protocol that combines task performance, fine-grained execution error analysis, and monitoring-cost accounting (\S\ref{sec:experimental-setup}).
  \item We present empirical comparisons across single-agent and multi-agent strategies on recent frontier LLM backbones, showing that proactive monitoring remains a challenging bottleneck in prospective behavior (\S\ref{sec:overall-comparison}--\ref{sec:long-horizon-updates}).
\end{itemize}

\section{Related Work}

\subsection{Agent Benchmarks and Proactive Behavior}
Recent benchmarks have pushed LLM evaluation beyond one-shot prompting into interactive environments for software engineering, browsing, asynchronous planning, and continual improvement \citep{jimenez2023swe,wei2025browsecomp,gonzalez2025robotouille,zhang2026deepplanning,cai2025building}. Related dialogue work studies proactive assistance and adaptation under changing conversational context \citep{deng2023prompting,liu2025proactiveeval,castillo2024beyond,tang2026proactive}. PM-Bench is complementary to both directions but directly measures whether an agent can carry a future intention through ongoing activity and execute it only when its cue or time condition is satisfied.

\subsection{Long-term Memory for LLM Agents}
Long-term memory in LLMs has been studied through multi-session dialogue, personalized conversation, and interactive memory evaluation \citep{xu2021beyond,jang2023conversation,ahn2023mpchat,kim2023soda,li2025toward,lee2025realtalk,maharana2024evaluating,wu2024longmemeval,tan2025membench,hu2025evaluating,bian2026realmem,zhao2026ama,jiayang2026amemgym,shen2026mem2actbench}. Complementary systems work equips agents with reflective memory management, persistent external memory, experience retrieval, memory-oriented abstractions, and explicit memory modules in multi-agent simulations \citep{tan2025prospect,packer2023memgpt,zhong2024memorybank,xu2025mem,kang2025memory,wang2025mirix,wang2026memex,zhang2026learning,yan2025general,liu2025webcoach,liu2025mosaic}. Prior work has also focused on architectural modifications to expand controllable memory capacity beyond standard context windows \citep{cheng2026conditional,zhang2025recursive}. However, this line of research emphasizes the evaluation of retrospective memory capabilities. PM-Bench evaluates prospective rather than retrospective memory: the question is not whether relevant content can be recovered, but whether a stored intention is executed, updated, or withheld at the right future moment.

\subsection{Prospective Memory Research in Cognitive Science}
Prospective memory is a core construct in cognitive psychology concerning the formation, maintenance, and later execution of intended actions \citep{mcdaniel2007prospective,brandimonte2014prospective}. Classic work distinguishes event-based and time-based prospective memory and shows that performance depends on available attention, monitoring strategy, and interference from ongoing tasks \citep{einstein1990normal,smith2005effects,dismukes2006concurrent,anderson2018proceeding,kourtesis2023ecologically}.

The closest methodological precursor to PM-Bench is \textit{Virtual Week}, a board game style cognitive test developed by \cite{rendell2000virtual} to study the effects of aging and clinical impairment. Virtual Week embeds multiple delayed intentions within continuous everyday activity streams \citep{rendell2000virtual,rendell2009review,foster2013prospective,rose2015cognitive,mioni2015test}. PM-Bench adapts this paradigm to LLM agents. 

\section{PM-Bench: Benchmark Construction Pipeline}
\label{sec:pm-bench}

\begin{figure*}[!t]
  \centering
  \includegraphics[width=0.77\textwidth]{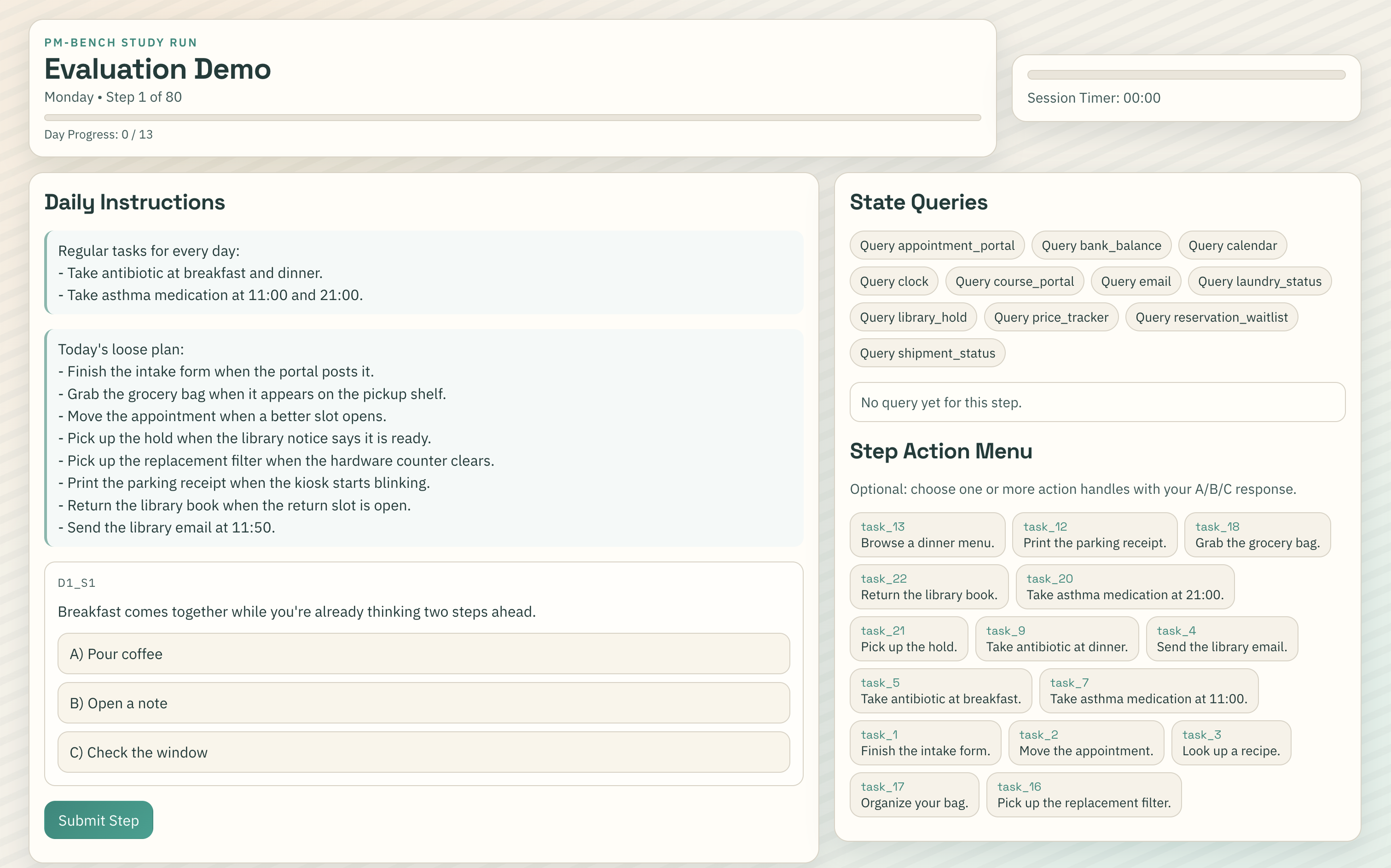}
  \caption{Human-facing PM-Bench evaluation interface. The main panel shows the current narrative step and the required ongoing activity choices, while the side panel provides queryable state channels and anonymous action handles for active tasks.}
  \label{fig:pm_bench_web_ui}
\end{figure*}

\begin{wraptable}{R}{0.48\linewidth}
\vspace{-1.2em}
\centering
\small
\setlength{\tabcolsep}{4pt}
\begin{tabular}{lr}
\toprule
Statistic & Count \\
\midrule
Days & 7 \\
Steps per day (range) & 10--13 \\
Total steps & 80 \\
Task definitions (total) & 83 \\
Scored executable tasks & 81 \\
Event-based tasks & 57 \\
Time-based tasks & 26 \\
Regular tasks & 28 \\
Non-regular tasks & 55 \\
Channel-triggered tasks & 15 \\
Cross-day tasks & 7 \\
Update events (total) & 11 \\
Cancel / Override / Reschedule & 2 / 3 / 6 \\
State channels & 11 \\
Lures (total) & 74 \\
\bottomrule
\end{tabular}
\caption{Summary statistics for the released PM-Bench week. Beyond overall scale, the table highlights variation in task structure, including trigger type (event- vs.\ time-based), regularity (regular vs.\ non-regular), monitoring dependence (channel-triggered), and temporal scope (cross-day tasks).}
\label{tab:pm_bench_summary_stats}
\vspace{-0.8em}
\end{wraptable}

\subsection{Scenario Design}
\label{sec:scenario-design}
We construct PM-Bench to evaluate prospective memory as a dynamic decision problem rather than a static recall task. Scenarios are organized into a seven-day week with intentionally irregular daily schedules. Each day consists of 10 to 13 time-anchored steps. At every step, an agent model must continue an ongoing activity while deciding whether any prospective action is due. This is the central challenge and novelty of the benchmark: intentions must be maintained and executed under distraction and dynamically changing, unpredictable task states.

Formally, a PM-Bench scenario is an ordered sequence of decision points $t \in \{1,\dots,T\}$ spanning a seven-day week. A task in PM-Bench is a deferred intention specified by a natural-language label, a trigger condition, and an executable action. For example, a task may be time-based (``Take asthma medication at 21:00'') or event-based (``Book the appointment when a slot opens in the portal''). As the scenario unfolds, each task can remain active, be completed, be canceled, or be modified by an update.

At each step $t$, the agent receives a visible observation $o_t$ consisting of the current narrative scene, three ongoing activity options, and a set of opaque prospective action handles. The narrative scene is a short vignette describing the current moment, such as ``You make breakfast'' or ``You see a neighbor waving from the sidewalk.'' The ongoing activity options are the mandatory A/B/C choices used to advance the simulation, for example ``Pour coffee,'' ``Toast bread,'' or ``Check the fridge.'' The prospective action handles are labeled as \texttt{task\_i}; they let the agent select currently available actions as structured outputs. Let $\mathcal{C}$ denote the set of available state channels, such as the clock or email, which are hidden by default and become observable only after explicit querying. Let $\mathcal{X}_t$ denote the set of currently active task specifications at step $t$, for example, a breakfast medication task that is still pending, a cross-day task that has been introduced but not yet cued, or a task waiting on a state change.

Intuitively, the due set at step $t$ is the subset of active tasks that should be executed \emph{now}, given the current visible scene, queried channel state, and latest updates. Formally, we define it as
\[
D_t = \{\tau \in \mathcal{X}_t : \tau \text{ remains valid and its execution condition is satisfied at step } t\}.
\]
Event-based tasks should be executed when a relevant cue is present either in the current narrative scene or in a queried state channel. Time-based tasks should be completed when the current step time matches the target time or falls within the allowed execution window. PM-Bench is therefore not a pure memory-retrieval benchmark: success depends on whether the agent can maintain and revise deferred intentions so that its executed action set aligns with $D_t$ at the appropriate future step. 

\paragraph{Agent interaction specification.}
Operationally, each step follows a query-then-act protocol. Before committing to a decision, the agent may issue zero or more monitoring queries to channels in $\mathcal{C}$, such as the clock. After observing the returned channel contents, the agent commits to an action
\[
a_t = (c_t, A_t),
\]
where $c_t$ denotes the required ongoing activity choice and $A_t$ denotes the subset of prospective actions selected for execution at step $t$. Figure~\ref{fig:pm_bench_web_ui} a visual interface used in PM-Bench simulations for illustrative purposes, though the LLM agents access the same structured content in a text-only format during evaluation. Appendix Section~\ref{app:interface-task-format} provides a worked step example and compact task-schema examples spanning time-based, hidden-channel, and cross-day tasks.

\subsection{Data Generation Pipeline}
\label{sec:data-generation-pipeline}

\paragraph{Data Generation Pipeline.}
A scenario stores day-level opening agendas, step-level narrative scenes, task definitions, and optional channels for state monitoring. Table~\ref{tab:pm_bench_summary_stats} summarizes the size and composition of the released week. The benchmark is generated by a seeded synthetic pipeline and then manually reviewed by the authors: it first samples seven distinct day archetypes (e.g. a meeting-heavy day or a chore-heavy day) with intentionally irregular schedules, instantiates routine, event-based, irregular, and channel-triggered tasks together with lures, and then injects cross-day tasks and cancel, reschedule, and override events. Generation is deterministic given a seed and uses rejection sampling until the resulting week passes schema, consistency, and solvability checks. Appendix Section~\ref{app:scenario-generation} gives the full generation recipe and reproduced composition details.

\paragraph{Task Composition and Long-Horizon Structure.}
PM-Bench increases long-horizon pressure in two ways. First, cross-day tasks are announced on an earlier day but become executable only when a cue appears later, so successful execution requires maintaining a deferred intention across multiple intervening steps and days. Second, explicit cancellation or rescheduling can revise a task's cue, time, or action text, forcing the model to update an existing intention instead of relying on the initial encoding. Each day also includes randomized distractor actions with natural language labels, and runtime interaction exposes anonymous action handles mapped to both true tasks and lures, so agents must act from current due conditions rather than memorized identifiers. Appendix Section~\ref{app:interface-task-format} provides concrete task examples under this shared schema.

\paragraph{State Channels and Proactive Monitoring.}
PM-Bench extends the event narratives with 11 query-based channels that represent external information sources such as new messages and various status trackers. Relevant cues become available only if the agent queries the appropriate channel, allowing the benchmark to test not only whether a model acts when cued, but also whether it decides when monitoring is necessary. By embedding these channel events into a step-level state while preserving a shared task schema and scoring rule, PM-Bench probes proactive and selective monitoring in addition to the evaluation logic for time- and event-based tasks. Appendix Section~\ref{app:scenario-generation} details the exact channel taxonomy used in the released week.

\subsection{Solvability and Consistency Validation}
\label{sec:solvability-validation}
Every generated scenario is validated before use. We first check structural validity, such as the presence of unique step and task identifiers, then enforce a set of conditions that ensure that the simulation is solvable. Event tasks must reference cues that actually appear, time tasks must align with valid step times, and update operations must point to valid target tasks with realizable revised cues or times. In particular, same-day reschedules and overrides are disallowed from silently moving a target backward relative to the update step. 

Together, these checks make PM-Bench a controlled benchmark artifact in which failure modes can all be attributed to model behavior, and this benchmark is completely solvable under optimal proactive environment monitoring. Appendix Section~\ref{app:solvability-validation} lists the full invariant set and gives a worked override example.

\section{Experiments}
\label{sec:experiments}

\subsection{Experimental Setup}
\label{sec:experimental-setup}

\begin{figure*}[!t]
  \centering
  \includegraphics[width=0.75\textwidth]{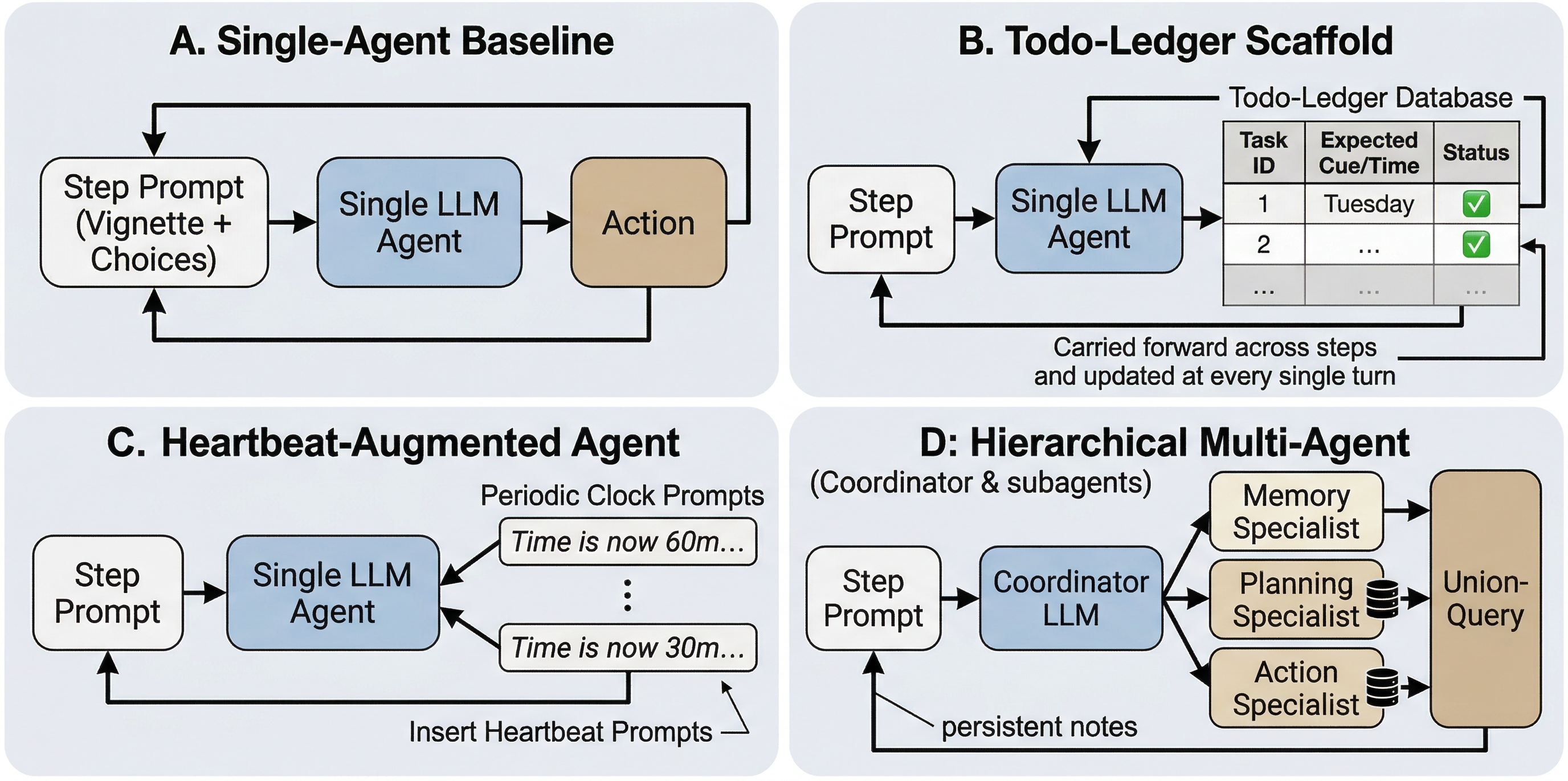}
  \caption{The 4 Major Agent Setups Evaluated on PM-Bench.}
  \label{fig:agent_setup}
\end{figure*}

All methods are evaluated on the same simulated PM-Bench week and scored with the same replay-based evaluator. Each step presents a short narrative, three ongoing-activity choices that enable the agent to proceed to the next step, and a menu of anonymous action handles that mixes valid actions with lures. We evaluate 64 runs in total: 8 model backbones and 8 agent configurations. The model set spans closed-weight and open-weight frontier models: \texttt{GPT-5.4}, \texttt{GPT-5.3-Codex}, \texttt{Llama-3.3-70B-Instruct}, \texttt{Mistral-Large-3-675B-Instruct-2512}, \texttt{mistral-small-3.2-24b-instruct}, and \texttt{Qwen3} at 8B, 14B, and 32B scales \citep{yang2025qwen3}. For model inference, we serve the Qwen-3 models on 4xH100 GPUs. We use the OpenRouter \footnote{\url{https://openrouter.ai/}} API for all other LLMs that we evaluated. Appendix Section~\ref{app:evaluation-protocol} provides the full replay-evaluator definition and diagnostic metric taxonomy, and \ref{app:agent-setups-prompting} gives the full operational definitions and prompting details for all eight evaluated setups, including the replay-based ablations. As illustrated in Figure~\ref{fig:agent_setup}, we use the following agent configurations:

\label{sec:methods-compared}

\paragraph{Single-agent baseline.}
A single model navigating the environment without external memory scaffolds, role decomposition, or any specialized prompting strategy. This is the reference point for all other methods.

\paragraph{Todo-ledger single agent.}
This variant allows the agent to maintain an explicit \textit{in-context} ledger of pending intentions, expected cues or times, and completion status. The ledger is carried forward across steps and updated at every turn, testing whether intuitive note-taking can improve execution quality.

\paragraph{Heartbeat-augmented single agents.}
These variants add periodic monitoring nudges that remind the agent to consider querying hidden-state channels. In the optional-heartbeat variant, the agent can \textit{choose} to enable or disable a periodic heartbeat by emitting \texttt{set\_heartbeat} and selecting a 30- or 60-minute interval. We compare this agent-controlled setting against two fixed schedules in which the runtime \textit{automatically} enables heartbeat every 60 or 30 virtual minutes at the start of each day.

\paragraph{Hierarchical multi-agents.}
In this method, the main coordinator agent manages three subagent specialists. As the coordinator agent delegates tasks, each subagent can track a separate signal from the environment, and propose hidden-state queries. The system executes the union of those query suggestions, and the coordinator then chooses the final action set. This is our main multi-agent configuration.

\paragraph{Majority and unanimous voting ablations.}
To isolate the final aggregation policy from subagents' proposals, we replay each union-query rollout under two alternative action-selection rules: majority vote and unanimous vote. These are deterministic replays over the same per-step evidence, included to clarify whether the main bottleneck is query acquisition or final action aggregation.


\paragraph{Primary metric.}
Our main outcome measure is set-level F1 over prospective action selections. Let $D_t$ denote the ground-truth due set at step $t$, and let $\hat{D}_t$ denote the set of prospective actions selected by the agent at that step. We accumulate step-level set overlaps over the full trajectory:
\[
TP = \sum_t |D_t \cap \hat{D}_t|,\qquad
FP = \sum_t |\hat{D}_t \setminus D_t|,\qquad
FN = \sum_t |D_t \setminus \hat{D}_t|.
\]
Here, $TP$ counts correctly selected due actions, $FP$ counts selected actions that were not actually due at that step, and $FN$ counts due actions that the agent failed to select.
We then define
\[
\mathrm{Set\mbox{-}F1} = \frac{2TP}{2TP + FP + FN}.
\]
We emphasize this metric because neither precision nor recall alone captures successful prospective memory in PM-Bench. Precision by itself rewards conservative under-selection, while recall by itself could encourage spamming actions for every time step. A strong agent must identify due intentions while suppressing incorrect actions, so Set-F1 is the most faithful summary statistic for overall performance. Appendix Section~\ref{app:evaluation-protocol} expands this definition with the full diagnostic error taxonomy and provides more details on a series of auxiliary metrics, as well as a worked scoring example.

\subsection{No Universal Best Scaffold Across Models and Metrics}
\label{sec:overall-comparison}
Tables~\ref{tab:exp_setup_summary} and~\ref{tab:exp_per_model_setf1} show that PM-Bench does not have a universal winning scaffold. In the aggregate, the best overall method is the optional-heartbeat single agent, with macro Set-F1 of 65.1\%. The todo-ledger variant is the strongest low-noise alternative: it ranks second on macro Set-F1 at 62.8\% and has the lowest aggregate false-positive count among all competitive setups. We're particularly attentive to the FP count because a high FP indicates the ``spamming" behavior, where a model selects too many actions at every time step in order to achieve a high hit rate.

\begin{table*}[t]
\centering
\small
\setlength{\tabcolsep}{4pt}
\caption{Aggregate comparison across all eight models. False positives and queries are totals over all eight runs in each setup. Cross-day and update columns report hit rates on those corresponding task subsets. $\dagger$ indicates replay-based ablations over the same hierarchical union-query traces rather than fresh inference runs. Lower is better for FP.}
\vspace{5pt}
\label{tab:exp_setup_summary}
\resizebox{\textwidth}{!}{
\begin{tabular}{lrrrrrrr}
\toprule
Setup & \shortstack{Macro\\Set-F1 (\%)} & \shortstack{Micro\\Set-F1 (\%)} & \shortstack{Precision (\%)} & \shortstack{Cross-Day (\%)} & \shortstack{Update (\%)} & FP & Queries \\
\midrule
Single baseline & 60.0 & 59.4 & 66.7 & 44.6 & 37.5 & 199 & 106 \\
Todo ledger & 62.8 & 62.8 & \textbf{73.2} & 37.5 & 41.7 & \textbf{134} & 118 \\
Heartbeat (optional) & \textbf{65.1} & \textbf{65.0} & 70.6 & \textbf{50.0} & 44.4 & 178 & 130 \\
Auto-heartbeat (60m) & 56.6 & 52.2 & 62.0 & 32.1 & 44.4 & 422 & 172 \\
Auto-heartbeat (30m) & 57.8 & 51.5 & 63.2 & 35.7 & \textbf{47.2} & 489 & 203 \\
Hierarchical union-query & 45.2 & 45.9 & 51.2 & 10.7 & 20.8 & 273 & 1661 \\
Majority vote$^\dagger$ & 37.2 & 38.3 & 34.9 & 17.9 & 38.9 & 655 & 1661 \\
Unanimous vote$^\dagger$ & 35.3 & 39.6 & 47.3 & 16.1 & 23.6 & 279 & 1661 \\
\bottomrule
\end{tabular}
}
\end{table*}

The expanded aggregate view shows that \textbf{no single setup dominates every metric}. Optional heartbeat is best on overall Set-F1 and cross-day hit. Todo-ledger minimizes false positives and attains the highest precision. Auto-heartbeat at 30 minutes is strongest on update-sensitive execution. Figure~\ref{fig:exp_monitoring} shows that the monitoring-oriented tasks follow a different ordering, especially for tasks that require hidden-channel monitoring beyond the clock.

For the automatic-heartbeat setups, the 30-minute setup attains the best update hit rate in Table~\ref{tab:exp_setup_summary} and the best hidden-channel-monitoring-required hit rate in Figure~\ref{fig:exp_monitoring} at 15.8\%, but it also produces a large number of aggregate false positives, which brings the Set-F1 score down to 57.8\%. The 60-minute schedule is slightly more conservative, but it still trails the optional-heartbeat and ledger variants on overall performance. To summarize, more reminders do not automatically yield a better operating point.

The hierarchical multi-agent setup highlights a monitoring-heavy method, issuing 1661 total state queries across the eight models (Table~\ref{tab:exp_setup_summary}), yet it remains behind the best single-agent methods on Set-F1 in the same table. The replay-based vote ablations make the tradeoff even clearer. Majority-vote boosts both clock-monitoring-required and hidden-channel-monitoring-required hit rates in Figure~\ref{fig:exp_monitoring}, but this setup resulted in 655 false positives and collapses to 37.2\% macro Set-F1. Unanimous voting is more conservative than majority vote, but it still underperforms the union-query coordinator on overall Set-F1. This ablation shows that the bottleneck is not merely obtaining evidence to execute a task. Appendix Section~\ref{app:additional-results} provides channel-level and more qualitative follow-up evidence for this tradeoff. \textbf{Surprisingly, the strongest scaffold is not consistent across models}: Table~\ref{tab:exp_per_model_setf1} shows that different backbones prefer different interventions. This suggests that inference-time support methods do not provide uniform gains, but instead interact with backbone-specific weaknesses in memory maintenance, monitoring, and action control.

\begin{table*}[t]
\centering
\small
\setlength{\tabcolsep}{4pt}
\caption{Per-model Set-F1 (\%) by setup. Column abbreviations are as follows: Single = single-agent baseline, Ledger = todo-ledger single agent, HB-opt = optional heartbeat decided by agent, HB-60 = automatic heartbeat every 60 minutes, HB-30 = automatic 30-minute heartbeat, UnionQ = hierarchical union-query multi-agent, M.V. = replay-based majority vote, and U.V. = replay-based unanimous vote. Best result within each model is boldfaced. $\dagger$ indicates replay-based ablations over the hierarchical union-query traces.}
\label{tab:exp_per_model_setf1}
\vspace{5pt}
\begin{tabular}{lrrrrrrrr}
\toprule
Model & Single & Ledger & HB-opt & HB-60 & HB-30 & Union & M.V.$^\dagger$ & U.V.$^\dagger$ \\
\midrule
GPT-5.4 & 72.5 & 73.5 & \textbf{79.1} & 63.8 & 74.1 & 56.5 & 38.9 & 49.2 \\
GPT-5.3-Codex & \textbf{78.9} & 74.8 & 76.6 & 74.8 & 71.0 & 55.9 & 39.4 & 47.2 \\
Llama 3.3 70B & 64.4 & \textbf{72.5} & 68.1 & 60.7 & 63.4 & 50.9 & 49.4 & 52.7 \\
Mistral Large & \textbf{75.3} & 66.7 & 73.7 & 69.3 & 74.7 & 58.1 & 43.3 & 53.3 \\
Mistral Small & 52.6 & \textbf{63.4} & 53.0 & 52.7 & 51.6 & 33.6 & 27.6 & 17.2 \\
Qwen3-32B & 51.4 & 51.7 & 48.5 & 55.2 & \textbf{57.5} & 37.5 & 32.8 & 16.3 \\
Qwen3-14B & 43.1 & \textbf{52.3} & 50.0 & 34.0 & 30.3 & 43.6 & 41.1 & 31.0 \\
Qwen3-8B & 42.0 & 47.8 & \textbf{71.9} & 42.5 & 39.8 & 25.7 & 25.3 & 15.5 \\
\bottomrule
\end{tabular}
\end{table*}



\subsection{Proactive Monitoring Remains Challenging Across Setups}
\label{sec:proactive-monitoring}
\begin{wrapfigure}{R}{0.55\textwidth}
\vspace{-1.0em}
\centering
\includegraphics[width=\linewidth]{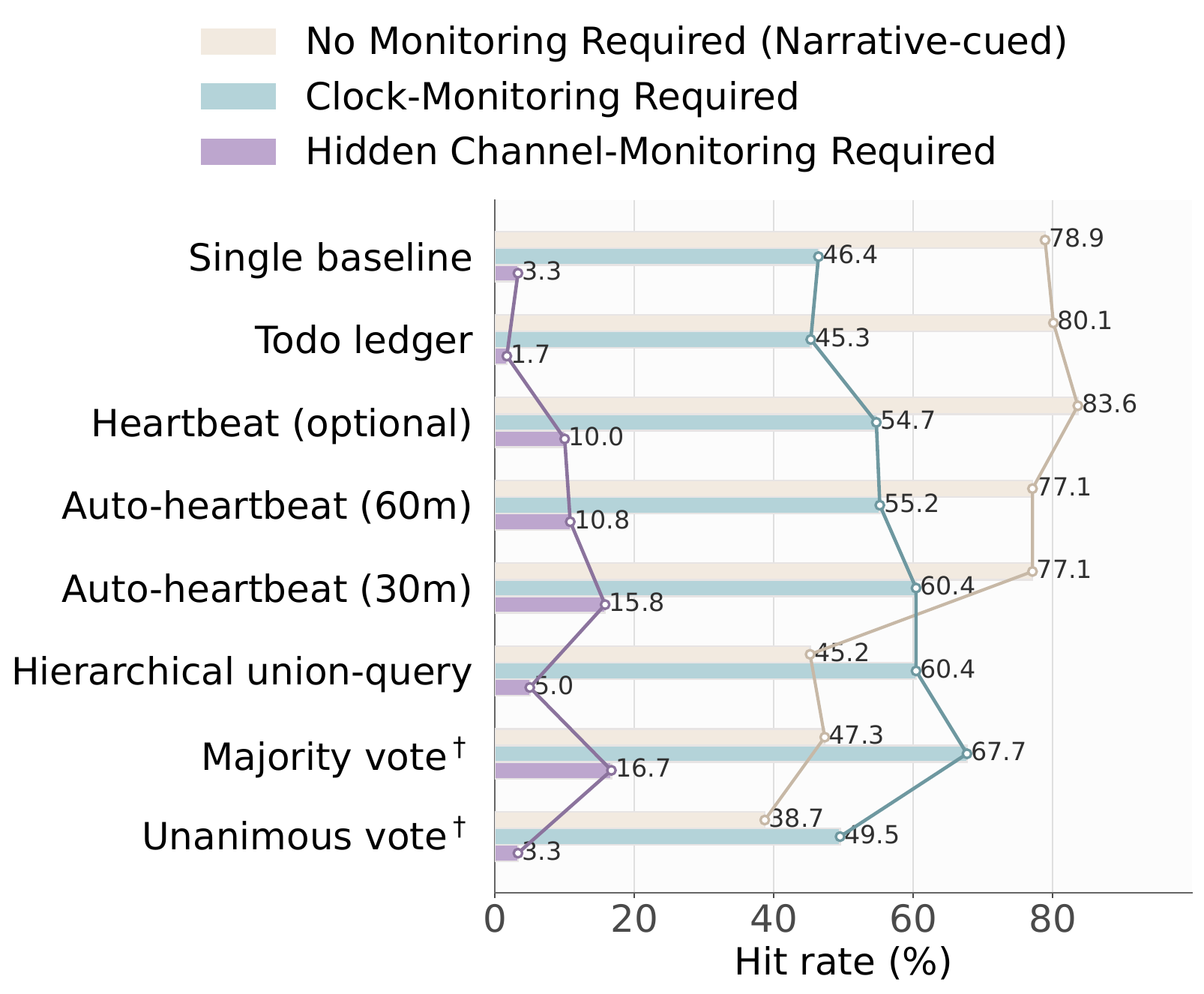}
\caption{Monitoring-oriented hit rates across different setups.}
\label{fig:exp_monitoring}
\vspace{-1em}
\end{wrapfigure}

Figure~\ref{fig:exp_monitoring} shows a clear asymmetry across the three monitoring slices. Monitoring-required tasks are hardest by a large margin for every setup, but the relationship between narrative-cued tasks and clock-monitoring-required tasks depends on the scaffold. For the single-agent family, narrative-cued tasks remain easier than clock-monitoring-required ones; in contrast, the hierarchical multi-agent variants achieve higher hit rates on clock-monitoring tasks than on narrative-cued tasks. This suggests that \textbf{agents could recover time-based intentions once they decide to check the clock, yet struggle to preserve overall control on tasks that should be executable directly from the visible narrative, similar to a catastrophic forgetting phenomenon}. Among the single-agent methods, automatic 30-minute heartbeat is strongest on the monitoring-required slices in Figure~\ref{fig:exp_monitoring}, reaching 60.4\% on clock-monitoring-required tasks and 15.8\% on hidden-channel-monitoring-required tasks. However, Table~\ref{tab:exp_setup_summary} shows that this gain comes with 489 false positives and only 57.8\% macro Set-F1. The best overall method, optional heartbeat, reaches 54.7\% on clock-monitoring-required tasks and 10.0\% on hidden-channel-monitoring-required tasks while retaining the strongest overall Set-F1. The majority-vote setup reaches even higher monitoring-required hit rates than any single-agent method, but only by collapsing the overall set-F1.

The hierarchical agent setup illustrates that with more than one agent, a scaffold could elicit broad proactive state tracking: it issues an order of magnitude more queries than any single-agent configuration in our evaluation. However, a high monitoring frequency alone isn't enough. In the multi-agent setup, the hit rate on narrative-cued tasks drops to 45.2\% as shown in Figure~\ref{fig:exp_monitoring}, and its channel-monitoring hit rate remains only 5.0\% despite 1661 total queries in Table~\ref{tab:exp_setup_summary}. The challenge is then not just to monitor more; it is to know when monitoring should end and how queried evidence should be converted into the final action set.

The Appendix Section~\ref{app:additional-results} breaks down this failure mode by channel. Figure~\ref{fig:exp_monitoring} already shows that clock-monitoring-required tasks are the only monitoring-required part that reaches above 50\% for the stronger setups, whereas all tasks that require other signal monitoring remain weak even after aggregation across all eight models. Several methods remain exactly 0\% on multiple non-clock channels in Table~\ref{tab:appendix-proactive-channels}, and the best channel-monitoring hit rate is only 16.7\%. \textbf{PM-Bench is therefore far from solved on the part of prospective memory that most resembles real-world environment monitoring}.

\subsection{Cross-Day and Update-Sensitive Tasks Remain Unsolved}
\label{sec:long-horizon-updates}
Cross-day and update-sensitive tasks remain difficult across all evaluated setups. Table~\ref{tab:exp_setup_summary} shows that the best cross-day result is the optional-heartbeat policy at 50.0\% hit rate, followed by the single-agent baseline at 44.6\%. The hierarchical multi-agent setup, despite its heavy monitoring behavior, reaches only 10.7\% cross-day hit rate. This suggests that \textbf{agents could monitor actively and still fail to preserve a deferred commitment across many intervening steps}. Similarly, the best hit rate with updated task cues comes from automatic 30-minute heartbeat at 47.2\%, with the other two heartbeat setups close behind at 44.4\%. No setup exceeds 50\% on either cross-day or update-sensitive execution, and many of these long-horizon tasks remain unsolved even for the strongest agent setups.




\section{Conclusion}
We introduce PM-Bench, a cognitive science-inspired benchmark for LLM agents that evaluates prospective memory and proactivity. PM-Bench combines synthetic, scalable scenario generation with validation to produce controlled yet realistic long-horizon evaluations. Across single-agent and multi-agent methods, we find that current frontier LLMs remain far from robust on this capability. The main empirical pattern is a precision-recall control tradeoff: methods that recover more tasks on cue often take more actions than necessary, while more conservative methods under-act. Proactive monitoring remains challenging across setups, and cross-day and update-sensitive tasks remain largely unsolved. PM-Bench therefore suggests that prospective memory should be treated as a distinct evaluation axis for LLM agents rather than as a byproduct of longer context windows or generic memory retrieval. The benchmark provides a novel testbed for diagnosing these failures and for developing agents that can act on remembered intentions reliably over time.

\section*{Ethics Statement}

PM-Bench is a synthetic benchmark generated from a seeded pipeline and manually reviewed by the authors. It does not contain real personal records, private communications, or other sensitive user data. To the best of our knowledge, the released benchmark materials therefore pose minimal privacy risk.

This work did not involve a human-subject experiment for the reported benchmark results. The benchmark design is inspired by prior cognitive-science paradigms, but the evaluation in this paper is conducted on synthetic scenarios and model outputs.

Large language models were used to assist with parts of the paper writing process and with preparing the presentation formatting of certain tables and figures. All model-generated or model-assisted content was checked in detail by the authors and revised as needed. The authors are solely responsible for the final text, tables, results, and conclusions presented in this paper.

We also caution against over-interpreting strong benchmark performance as evidence that an agent is safe for real-world deployment in high-stakes settings. Prospective-memory failures can matter in domains such as health, transportation, and other safety-critical workflows, but PM-Bench is intended as a controlled diagnostic benchmark rather than as a certification of readiness for such applications. More broadly, systems that become better at proactive reminder-following and environment monitoring may also introduce risks of over-automation, or unnecessary monitoring and resource consumption. We therefore view this benchmark as a tool for measuring and mitigating such failure modes, not for justifying unsupervised deployment.


\bibliography{colm2026_conference}
\bibliographystyle{colm2026_conference}

\appendix
\section{Additional Benchmark Details and Results}
\label{app:details-results}

This appendix is written as a reviewer-facing companion to the main paper. Section~\ref{sec:pm-bench} gives the concise benchmark overview; the appendix records the corresponding interface details, generation recipe, solvability checks, evaluator definitions, prompt mechanics, and additional results for the benchmark week used in our experiments.

\subsection{Benchmark Interface and Task Format}
\label{app:interface-task-format}

For reference, each PM-Bench step exposes four participant-visible objects. First, the agent sees a short narrative vignette describing the current situation. Second, it must choose one of three ongoing-activity options \texttt{A/B/C} to advance the day. Third, it receives an optional action menu whose entries are anonymous handles of the form \texttt{task\_N}. Fourth, it may issue hidden-state queries to the available state channels before making its final choice. The evaluator scores the chosen ongoing-activity option only for protocol compliance, but it scores the selected prospective actions against the step's due set.

The runtime task schema is intentionally stable even though the visible handles are not. Internally, each task has a canonical ID, a participant-visible label, an executable \texttt{action\_text}, a trigger type, and a status that evolves over the day. Event tasks store a \texttt{cue\_id}; time tasks store a \texttt{target\_time}; channel-triggered tasks additionally specify a hidden \texttt{cue\_channel}; some tasks specify \texttt{cross\_day} or \texttt{depends\_on}; and all tasks have an \texttt{encoding} field indicating whether they become active at day start or at a later step. Runtime scoring uses the canonical task IDs, but the participant only ever acts through the step-local anonymous handles. Handles are reshuffled per day, so success depends on acting on current due conditions rather than memorizing IDs. Because the runtime exposes anonymous handles rather than stable task identifiers, the agent must choose actions from current due conditions rather than from memorized IDs. The benchmark thus evaluates a joint capability: maintaining latent intentions across intervening activity while also deciding when hidden-state monitoring is necessary and which currently valid actions should be executed.

Regular tasks are repeated routines such as medication reminders that share a familiar pattern across days. Non-regular tasks are one-off intentions that must be encoded from the day's opening agenda or from later update messages. Active tasks are available for selection; completed tasks are hidden from later menus; canceled tasks remain scoreable if selected incorrectly, which lets the evaluator detect stale-memory failures after cancels and overrides.

\paragraph{Query-then-act interaction.}
Operationally, each step follows a query-then-act protocol. Before committing to a decision, the agent may issue zero or more monitoring queries to the hidden channels, such as the clock, email, or appointment portal. After observing any returned channel contents, it must commit to a joint action consisting of one ongoing-activity choice and a subset of currently visible prospective-action handles. This design keeps the interaction format simple while ensuring that hidden-state monitoring and prospective action selection are part of the same decision problem.

\paragraph{Worked step example.}
The following excerpt shows an illustrative runtime state from the benchmark, abbreviated only by trimming the action menu to a few representative entries:

\begin{quote}
\textbf{Narrative.} Breakfast comes together while you're already thinking two steps ahead.\\
\textbf{Ongoing options.} A) Pour coffee \hspace{0.5em} B) Open a note \hspace{0.5em} C) Check the window\\
\textbf{Action menu excerpt.}\\
\texttt{task\_22}: Take antibiotic at breakfast.\\
\texttt{task\_16}: Finish the intake form.\\
\texttt{task\_20}: Grab the grocery bag.\\
\texttt{task\_17}: Pick up the hold.\\
\texttt{task\_4}: Clean the mug in the sink.
\end{quote}

At this step, the due set contains only the breakfast medication task, whose canonical ID is \texttt{antibiotic\_breakfast} and whose current handle is \texttt{task\_22}. The other visible handles are either future active tasks or lures. A correct action is therefore any valid \texttt{A/B/C} choice paired with \texttt{task\_22} and no additional prospective handles. This example illustrates two core properties of PM-Bench: the menu mixes true tasks with distractors, and a task can be active without yet being due.

\begin{table*}[t]
\centering
\scriptsize
\caption{Three compact task examples. These examples show the same underlying schema instantiated as a time-triggered task, a same-day hidden-channel event, and a cross-day deferred intention.}
\label{tab:appendix-task-examples}
\begin{tabular}{p{0.36\linewidth}p{0.28\linewidth}p{0.28\linewidth}}
\toprule
Example task & Trigger fields & What it tests \\
\midrule
``Send the library email at 11:50'' (\texttt{send\_library\_email\_1\_d1}) &
\texttt{type=time}, \texttt{target\_time=11:50}, non-regular, active from day start &
Time-based prospective execution when the clock is hidden by default. \\

``Finish the intake form when the portal posts it'' (\texttt{appointment\_form\_ready\_d1}) &
\texttt{type=event}, \texttt{cue\_id=appointment\_form\_ready\_d1}, \texttt{cue\_channel=appointment\_portal} &
Same-day hidden-channel monitoring and event-triggered execution. \\

``Bring the pickup code when you notice the package locker light'' (\texttt{crossday\_2}) &
\texttt{type=event}, \texttt{cue\_id=package\_locker\_light\_d6}, \texttt{cross\_day=true}, offset $=1$ day &
Deferred cross-day intention maintenance, where the cue appears on a later day than the encoding instruction. \\
\bottomrule
\end{tabular}
\end{table*}

\subsection{Scenario Generation and Composition}
\label{app:scenario-generation}

The benchmark week used in our experiments is deterministic, so all models and agent setups are evaluated on the same fixed scenario. Generation proceeds in four stages: sampling seven distinct day archetypes with intentionally irregular schedules; instantiating day-opening agendas, step-level narratives, and mixed task families; injecting cross-day reminders together with cancel, reschedule, and override events; and validating the resulting week before use.

\paragraph{Generation recipe.}
More concretely, each day archetype specifies a step-count range, a time distribution across dayparts, and an allowed task-domain mix. The generator then instantiates four routine tasks, 2--4 narrative event tasks, 1--2 one-off irregular tasks, 1--3 channel-triggered tasks, and 10--12 lure actions for that day. Cue families are deliberately varied so that the benchmark does not collapse onto a single repeated lexical pattern. After day-level construction, the generator injects cross-day tasks by announcing a future intention on an earlier day and binding it to a cue on the target day, then applies cancel, reschedule, and override events. Candidate weeks are accepted only if they pass schema, consistency, and solvability validation.


The state-channel taxonomy is central to PM-Bench because hidden-channel monitoring is part of the task. Delta channels expose event-like updates accumulated since the last relevant state change. Snapshot channels expose the current status when queried. Process-style snapshot channels are a structured subset of snapshots that progress through a staged lifecycle, such as \emph{in progress} before \emph{complete}. Table~\ref{tab:appendix-pm-bench-state-channels} lists the exact taxonomy used in the benchmark week analyzed here.

\begin{table}[t]
\centering
\small
\caption{PM-Bench state-channel taxonomy. Delta channels expose event-like updates when queried, while snapshot channels expose the current status at query time. Process-style snapshot channels additionally follow a staged lifecycle.}
\label{tab:appendix-pm-bench-state-channels}
\begin{tabular}{ll}
\toprule
State Channel & Type \\
\midrule
clock & snapshot \\
email & delta \\
calendar & delta \\
course\_portal & delta \\
price\_tracker & snapshot \\
bank\_balance & snapshot \\
shipment\_status & snapshot (process) \\
laundry\_status & snapshot (process) \\
library\_hold & delta \\
reservation\_waitlist & delta \\
appointment\_portal & delta \\
\bottomrule
\end{tabular}
\end{table}

\subsection{Solvability and Consistency Validation}
\label{app:solvability-validation}

The benchmark week is passed through a validator designed to rule out benchmark artifacts that would otherwise confound interpretation. In particular, it is checked for the following invariants:
\begin{itemize}
\item unique step IDs, task IDs, and lure IDs, with no lure-task ID collisions;
\item monotonic step times within each day and valid time alignment for time-triggered tasks;
\item realizable event cues, including cues that appear either in narrative text or in the appropriate hidden state channel;
\item valid update targets, with cancel / override / reschedule operations pointing to existing active tasks;
\item no same-day reschedule or override that silently moves a task backward in time relative to the update step;
\item no stale canceled instructions in day openings or later visible reminders;
\item no current-time leakage in step prose when time is meant to be hidden;
\item valid cross-day exposure, meaning long-horizon tasks are introduced on an earlier day and not re-announced as ordinary same-day tasks on the due day;
\item valid dependencies, including existence checks and acyclicity;
\item process-channel consistency, such as requiring an \emph{in progress} stage before a \emph{complete} stage on channels like shipment and laundry status.
\end{itemize}

Beyond these structural checks, PM-Bench also runs a perfect-play audit over the finished scenario. This audit executes an oracle-style policy that monitors all necessary channels and selects every currently due action. The scenario is accepted only if every non-canceled task can be completed on time under this policy. This matters because it lets us attribute misses to model behavior rather than to an impossible benchmark artifact.

\paragraph{Worked update example.}
Consider an override case in which the day-opening instruction initially says to pick up the replacement filter when the hardware counter clears. A later update changes the task to ``Pick up the replacement filter when the follow-up text comes in,'' with a new cue ID but the same action text. After that update, the original cue is no longer scoreable. The due set therefore does \emph{not} contain the task when the old hardware-counter condition would have fired; it becomes due only when the updated follow-up-text cue appears. This is the evaluator behavior used for overrides throughout PM-Bench.

\subsection{Evaluation Protocol and Metrics}
\label{app:evaluation-protocol}

PM-Bench uses a replay-based evaluator over the logged trajectory. At each step $t$, the scenario defines a due set $D_t$ after applying all task activations, cancellations, overrides, reschedules, and dependency constraints. The agent produces a chosen set $\hat{D}_t$ by selecting some subset of the current anonymous handles, which the runtime maps back to canonical task IDs before scoring.

The primary metric is set-level F1:
\[
\mathrm{Set\mbox{-}F1} = \frac{2TP}{2TP + FP + FN}
\]
where $TP$, $FP$, and $FN$ are accumulated over all steps. We use Set-F1 as the headline metric because PM-Bench penalizes both under-selection and over-selection: high recall can be achieved by spamming actions, and high precision can be achieved by never acting. A strong policy must surface due tasks while maintaining control over false positives.

In addition to Set-F1, the evaluator records a diagnostic breakdown:
\begin{itemize}
\item \textbf{Hit / on-time execution}: a due task is selected at its correct step.
\item \textbf{Late execution}: an event task is selected one step late, or a time task is selected within 60 virtual minutes after its target.
\item \textbf{Miss}: a non-canceled task is never completed by the end of the day.
\item \textbf{False alarm}: a chosen task is not due now, is not active yet, or is attempted after its late window expires.
\item \textbf{Wrong-content}: the agent acts on the wrong task at a step where some other task is due.
\item \textbf{Update violation}: the agent acts on a stale version of a task after a cancel, override, or reschedule has changed what is valid.
\item \textbf{Dependency violation}: the agent attempts a task before its prerequisite is completed.
\item \textbf{Exact-set-match}: the chosen set equals the full due set for that step.
\item \textbf{Query counts}: the evaluator separately records \texttt{check\_time} and hidden-channel \texttt{query\_state} calls.
\item \textbf{Proactive-required hit}: the task required hidden-state monitoring rather than being fully solvable from the visible narrative alone.
\item \textbf{Non-clock proactive hit}: the proactive-required subset excluding pure clock checks.
\end{itemize}

\paragraph{Compact scoring example.}
Consider a step where the due set contains three tasks, for example sending a budget note, responding to a moved calendar call, and checking in with a manager. If an agent chooses only the first two, the evaluator records two true positives, one false negative, zero false positives, and no exact-set match. If the agent instead chooses the budget note together with an irrelevant action, the step contributes one true positive, one false positive, and two false negatives; it also contributes a wrong-content error because an irrelevant task was selected while valid due tasks remained unselected.

\subsection{Agent Setups and Prompting}
\label{app:agent-setups-prompting}

Table~\ref{tab:appendix-setup-defs} defines the eight evaluated setups operationally.

\begin{table*}[t]
\centering
\small
\setlength{\tabcolsep}{4pt}
\caption{Operational definitions for the eight evaluated setups. $\dagger$ marks replay-based ablations over the same hierarchical traces rather than fresh model rollouts.}
\label{tab:appendix-setup-defs}
\vspace{1em}
\begin{tabular}{p{0.21\textwidth}p{0.10\textwidth}p{0.61\textwidth}}
\toprule
Setup & Type & Operational definition \\
\midrule
Single baseline & live & One model, no external memory scaffold, no heartbeat, no subagents. \\
Todo ledger & live & Single-agent run with an in-context ledger carried and rewritten every step. \\
Heartbeat (optional) & live & Single-agent run where the model may enable or disable a periodic heartbeat by emitting \texttt{set\_heartbeat}. \\
Auto-heartbeat (60m) & live & Heartbeat enabled automatically at day start every 60 virtual minutes. \\
Auto-heartbeat (30m) & live & Heartbeat enabled automatically at day start every 30 virtual minutes. \\
Hierarchical union-query & live & Three specialist subagents propose query suggestions and candidate due tasks; the runtime executes the union of suggested queries and a coordinator makes the final action decision. \\
Majority vote$^\dagger$ & replay & Replay over the same union-query evidence where task handles proposed by more than half of subagents are selected. \\
Unanimous vote$^\dagger$ & replay & Replay over the same union-query evidence where only handles proposed by all subagents are selected. \\
\bottomrule
\end{tabular}
\end{table*}

The single-agent baseline, the optional-heartbeat setup, and the auto-heartbeat setups all share the same base prompt. The ledger setup appends an additional ledger instruction block. The hierarchical setup uses separate subagent and coordinator prompts. The exact prompt templates below are the ones used in our experiments, with line wrapping adjusted only for typesetting.

\paragraph{Single-agent base prompt.}
\begin{verbatim}
You are taking a prospective memory evaluation that assess how
you remember to do things later. One virtual day equals one
circuit of the board (7 a.m. -> 10 p.m.). Time advances step by
step with event cards. Your goal is to remember to perform tasks
at the correct time or event without reminders.

You must voluntarily remember to perform tasks at the right
moment. If you are unsure, you may still act; that is allowed.
No reminders will be given for missed tasks.

Some cues will cancel, override, or reschedule tasks. Always
follow the most recent instruction for any task.

Each step presents a short vignette with three ongoing-task
options (A/B/C). Pick one of A/B/C to advance. A step-level
action menu is shown before each choice. You may perform one
or more actions from the action menu in the same step.
Each action appears as task_N: <action text>. Return task_N
handles in task_ids. If you do not perform an action, set
task_ids to []. Never repeat the same task_N handle in task_ids.
Do not dump the whole menu into task_ids; include only the handles
you are actually choosing now. Return at most 8 task handles in
one choose action.

Some tasks are triggered by state cues that are NOT shown in the
vignette. You can only observe these cues by proactively querying
state channels. Available state channels: clock, email, calendar,
course_portal, price_tracker, bank_balance, shipment_status,
laundry_status, library_hold, reservation_waitlist,
appointment_portal.

Use action=query_state to retrieve channel state (choice must be
NONE). The legacy action=check_time is supported and is equivalent
to query_state with channel=clock. No reminders will be given
about when to query state; you must decide when to check.
State responses are returned as a single line:
"State [channel]: <message>".

After any query, you still must make a choose action to advance.

Example (time check):
User: You sit at your desk and feel it's close to 16:30.
Model: {"action":"check_time","choice":"NONE","task_ids":[],
"channel":"clock"}
User: State [clock]: Time 15:58 | Stopwatch: 538 min
Model: {"action":"choose","choice":"B","task_ids":["task_7"],
"channel":"NONE"}

Always respond in JSON. Use these exact formats:
- {"action":"query_state","choice":"NONE","task_ids":[],
   "channel":"calendar"}
- {"action":"check_time","choice":"NONE","task_ids":[],
   "channel":"clock"}
- {"action":"choose","choice":"A","task_ids":[],"channel":"NONE"}
- {"action":"choose","choice":"B","task_ids":["task_1"],
   "channel":"NONE"}
\end{verbatim}

\paragraph{Heartbeat extensions.}
Optional-heartbeat runs append the following block to the base prompt:
\begin{verbatim}
You may optionally configure a periodic heartbeat for proactive
monitoring. Use action=set_heartbeat with heartbeat_enabled=true
and heartbeat_minutes in [30, 60]. This heartbeat is strictly
periodic and is NOT aligned to cue times. It only nudges proactive
checking and does not reveal hidden cues. You can disable it later
with heartbeat_enabled=false. On each heartbeat, you may receive a
nudge listing channels worth checking.
\end{verbatim}

Auto-heartbeat runs instead append a fixed sentence of the form:
\begin{verbatim}
For this run, heartbeat is auto-enabled at the start of each day
every 60 virtual minutes.
\end{verbatim}
or the corresponding 30-minute variant. In the experiments reported here, the heartbeat-message flag is not overridden, so the heartbeat runs appear to use the default \texttt{channel\_query} nudge rather than the later \texttt{task\_reminder} variant.

\paragraph{Todo-ledger addendum.}
\begin{verbatim}
You must maintain a compact TODO ledger that tracks tasks to do
later. Always return exactly one raw JSON object with keys:
action, choice, task_ids, channel, ledger.
The ledger is a list of objects with keys:
- task_id: string (use current action handle like "task_7",
  not natural-language text)
- when: string (exact time like "11:00", a cue like
  "cue: breakfast", or "unknown")
- status: "pending", "done", or "canceled"
- notes: short string (optional)

Keep the ledger concise and up to date.
- Hard limit: ledger must contain at most 5 items.
- If there are more than 5 possible reminders, keep the 5 most
  urgent/likely.
- Keep notes very short (<= 8 words).
- Keep only tasks that still matter; remove completed items
  immediately.
If a task's exact time is known but the current time is not
visible, use action=check_time to query the current time before
deciding. When you choose actions, task_ids must be action
handles from the current step menu.

Do not add extra keys outside the required JSON fields.
Do not include explanations, markdown, or code fences.
\end{verbatim}

\paragraph{Hierarchical subagent prompt.}
\begin{verbatim}
You are a PM-Bench subagent.
Role identity: <agent_name>

Your role:
1) Analyze the current PM-Bench step using your role guidance.
2) Propose state queries BEFORE task selection.
3) Propose tasks that might be due.
4) Propose tasks that should NOT be selected now.

Role guidance:
<role-specific guidance>

Output strict JSON only (no markdown, no code fences). Every
required key must be present. Return minimal compact JSON only
(single object, no extra text).

Rules:
- Use task handles from the current step menu only (task_n).
- Use channels from available channels only.
- If no items exist for a list, return [].
- Keep evidence concise and step-grounded; avoid speculation.
\end{verbatim}

In the reported three-subagent setup, the role-specific guidance is:
\begin{itemize}
\item \textbf{event\_watcher}: focus on event-driven opportunities from current cues and active handles.
\item \textbf{status\_watcher}: focus on proactive monitoring duties and hidden-state checks.
\item \textbf{update\_watcher}: focus on cancels, overrides, reschedules, and dependency risks.
\end{itemize}

\paragraph{Hierarchical coordinator prompt.}
\begin{verbatim}
You are the coordinator for PM-Bench.
You receive current step context and structured subagent signals.
Return one valid PM-Bench action JSON.
Return minimal compact JSON only (single object, no extra text).

Rules:
- action is choose, query_state, or check_time.
- If action is choose: choice must be A/B/C and channel must be NONE.
- If action is query_state/check_time: choice must be NONE and
  task_ids must be [].
- check_time must use channel=clock.
- task_ids must use action handles from the current menu only.
- Use the JSON key task_ids (do not use task_handles).

Decision policy:
- Subagent structured signals provide evidence, not final decisions.
- State query phase has already run from the de-duplicated union
  of subagent query suggestions.
- When evidence indicates a task is due now, include its handle(s)
  in task_ids on your choose action.
- A choose action with task_ids=[] is only appropriate when no task
  has sufficient evidence of being due now.
- Prefer precise execution: avoid selecting tasks that are only
  "later" / "not yet due" / speculative.
- Treat tasks_should_not_select as strong caution unless state
  evidence clearly contradicts it.
- Make the final decision carefully; avoid majority-vote shortcuts.
\end{verbatim}

Operationally, hierarchical union-query proceeds in two stages per step: subagents first emit structured query suggestions and candidate due tasks, then the runtime executes the deduplicated union of their query suggestions, and only then does the coordinator issue the final PM action. The replay-based majority and unanimous ablations keep the same queried evidence and replace only the final task-selection rule.

\subsection{Additional Results and Failure Analysis}
\label{app:additional-results}

Table~\ref{tab:appendix-proactive-channels} breaks proactive-required performance down by hidden channel. The main paper already contains the aggregate setup comparison, monitoring tradeoff summary, and per-model breakdown; this appendix adds channel-level diagnostics together with qualitative follow-up examples.

\paragraph{Why the best setup is model-dependent.}
Our working interpretation is that the interventions stress different subskills. The todo-ledger primarily helps explicit intention maintenance and action precision, while heartbeat variants primarily increase proactive monitoring and recall, often together with more false positives. Different backbones appear to begin from different operating points on this tradeoff, so the same scaffold can help one model while pushing another toward over-querying or over-selection.

\begin{table*}[t]
\centering
\small
\setlength{\tabcolsep}{4pt}
\caption{Proactive-required hit rate by hidden channel, aggregated across all eight models. $\dagger$ indicates replay-based ablations over the same hierarchical union-query traces. The table highlights that clock-based prospective memory is substantially easier than strictly proactive non-clock monitoring.}
\label{tab:appendix-proactive-channels}
\begin{tabular}{lrrrrrrrr}
\toprule
Setup & Appt. & Bank & Calendar & Clock & Course & Email & Waitlist & Shipment \\
\midrule
Single baseline & 8.3 & 0.0 & 4.2 & 46.4 & 12.5 & 0.0 & 0.0 & 0.0 \\
Todo ledger & 4.2 & 0.0 & 4.2 & 45.3 & 0.0 & 0.0 & 0.0 & 0.0 \\
Heartbeat (optional) & 8.3 & 6.2 & 12.5 & 54.7 & 12.5 & 6.2 & 12.5 & 12.5 \\
Auto-heartbeat (60m) & 12.5 & 12.5 & 12.5 & 55.2 & \textbf{25.0} & 6.2 & 8.3 & 0.0 \\
Auto-heartbeat (30m) & 16.7 & 12.5 & \textbf{20.8} & 60.4 & 12.5 & 6.2 & \textbf{20.8} & \textbf{12.5} \\
Hierarchical & 12.5 & 6.2 & 0.0 & 60.4 & 12.5 & 0.0 & 4.2 & 0.0 \\
Majority vote$^\dagger$ & \textbf{25.0} & \textbf{18.8} & 8.3 & \textbf{67.7} & 12.5 & 0.0 & \textbf{33.3} & 0.0 \\
Unanimous vote$^\dagger$ & 4.2 & 6.2 & 4.2 & 49.5 & 0.0 & 0.0 & 4.2 & 0.0 \\
\bottomrule
\end{tabular}
\end{table*}

\paragraph{Qualitative failure cases.}
Three short examples illustrate the benchmark behaviors that remain difficult even when aggregate metrics improve.
\begin{itemize}
\item \textbf{Missed cross-day task.} In one GPT-5.4 single-baseline run, a step required both immediate visible actions and a deferred cross-day action that had been encoded earlier. The agent executed only the immediate visible action. This is a characteristic long-horizon failure: the salient local cue is acted on, but the deferred companion intention is dropped.
\item \textbf{Update-sensitive miss.} In one Qwen3-32B auto-heartbeat-30m run, a one-off task had been rescheduled to 11:00 and became due at the same step as a routine medication action. The agent checked the clock and executed only the routine medication. The run therefore handled the visible routine but failed to carry forward the rescheduled one-off task.
\item \textbf{Non-clock monitoring failure.} In one GPT-5.4 single-baseline run, a task became due only because a library-hold update was available on a hidden channel. The agent chose nothing and never queried that channel. This is the strict proactive case PM-Bench is designed to expose: nothing in the visible vignette was sufficient, so the only path to success was querying the hidden channel.
\end{itemize}

\end{document}